\documentclass[pmlr]{jmlr}

\usepackage{longtable}

\usepackage{booktabs}
\usepackage[load-configurations=version-1]{siunitx} 

\makeatletter
\def\set@curr@file#1{\def\@curr@file{#1}} 
\makeatother

\theorembodyfont{\upshape}
\theoremheaderfont{\scshape}
\theorempostheader{:}
\theoremsep{\newline}

\usepackage{enumitem}
\usepackage{algorithm}
\usepackage{graphics}
\usepackage{algorithmic}
\usepackage{multirow}
\usepackage{adjustbox}

\title[]{Small-Group Learning, with Application to Neural Architecture Search}

\author{\Name{Xuefeng Du}
       \Email{xuefengdu1@gmail.com} 
       \AND
       \Name{Pengtao Xie\textsuperscript{*}}
       \Email{p1xie@eng.ucsd.edu}\\
       \addr 
University of California San Diego
\AND
       }

\begin{document}

\maketitle

\begin{abstract}
In human learning, an effective learning methodology is small-group learning: a small group of students work together towards the same learning objective, where they express their understanding of a topic to their peers, compare their ideas, and help each other to trouble-shoot problems. In this paper, we aim to investigate whether this human learning method can be borrowed to train better machine learning models, by developing a novel ML framework -- small-group learning (SGL). In our framework, a group of learners (ML models) with different model architectures collaboratively help each other to learn by leveraging their complementary advantages.  Specifically, each learner uses its intermediately trained model to generate a pseudo-labeled dataset and re-trains its model using  pseudo-labeled datasets generated by other learners. SGL is formulated as a multi-level optimization framework consisting of three learning stages: each learner trains a model independently and uses this model to perform pseudo-labeling; each learner trains another model using datasets pseudo-labeled by other learners; learners improve their architectures by minimizing validation losses. An efficient algorithm is developed to solve the multi-level optimization problem. We apply SGL for neural architecture search. Results on CIFAR-100, CIFAR-10, and ImageNet demonstrate the effectiveness of our method.
\end{abstract}

\section{Introduction}

\let\thefootnote\relax\footnotetext{$^*$Corresponding author.}

Small-group learning is a broadly practiced learning methodology in humans' learning activities with high efficacy.  In SGL, a small group of students collaborate with each other to study the same topic. Each student conveys his/her understanding of this topic to others and deepens the understanding by referring to others' thoughts. 
SGL can effectively facilitate  more profound and meaningful learning.

\begin{figure}[t]
    \centering
 \includegraphics[width=0.5\columnwidth]{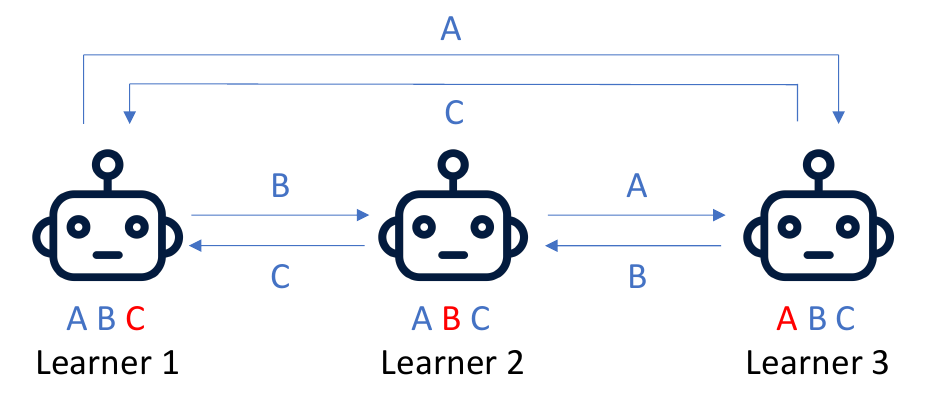}
       \caption{Illustration of small-group learning. There are three learners performing a classification task with three classes: A, B, and C. Blue letters denote classes that learners are good at (e.g., learner 1 performs well on class A and B) and red letters denote classes that learners are not good at. For each class, strong learners help weak learners to learn. For example, learner 2 and 3 are strong learners on class C; they help learner 1 (which is a weak learner on C) to improve the performance on C.}
 \label{fig:illus}
\end{figure}

We are interested in asking whether this human learning skill can be translated into a machine learning skill to train better ML models.  In a machine learning task (e.g., multi-class classification), it is typically the case that models with different architectures have complementary advantages. On a certain class $c$, model $A$ (e.g., ResNet~\citep{resnet}) performs better than model $B$ (e.g., DenseNet~\citep{HuangLMW17}). On another class $d$, model $B$ performs better than model $A$. It is desirable to let these two models help each other during learning so that both of them eventually achieve great performance on both classes: specifically, model $A$ helps model $B$ to improve its performance on class $c$ and model $B$ helps model $A$ to improve its performance on class $d$. Based on this idea, we propose a novel learning framework -- small-group learning (SGL) (as illustrated in Figure~\ref{fig:illus}). SGL involves a set of ML learners which solve the same learning task $T$. Without loss of generality, we assume $T$ is classification while noting that our framework can be broadly applied to other tasks as well. Each learner $k$ has a learnable neural architecture $A_k$ and two sets of network weights $W_k$ and $V_k$. The architectures and network weights of different learners are different. All learners share the same training dataset and validation dataset. These learners help each other to learn, in the following way: given an unlabeled dataset, each learner uses its intermediately trained model (including architecture and network weights) to generate pseudo-labels on this dataset and each learner leverages the pseudo-labeled datasets generated by other peers to re-train its model. On each class $c$, models performing well on $c$ can generate high-quality pseudo-labeled datasets w.r.t $c$. Trained on these datasets, models that originally do not perform well on $c$ can improve their performance on $c$. In this framework, there are three learning stages. In the first stage, each learner $k$ trains its network weights $V_k$ on the training dataset, with its architecture $A_k$ fixed. In the second stage, each learner $k$ uses its optimally trained $V^*_k$  in the first stage to make predictions on an unlabeled dataset and generates a pseudo-labeled dataset; then each learner $k$ trains its second set of network weights $W_k$ using the pseudo-labeled datasets generated by other learners and using a human-labeled training dataset. In the third stage, each learner $k$ updates its architecture $A_k$ by minimizing prediction losses on the validation dataset. A multi-level optimization framework is developed to organize the three learning stages and enable them to be performed in a joint manner. Our method is applied for neural architecture search tasks on CIFAR-100, CIFAR-10, and ImageNet~\citep{deng2009imagenet}. Experimental results demonstrate the broad effectiveness of SGL.

The major contributions of this paper are:
\begin{itemize}
\item Drawing inspiration from the small-group learning (SGL) methodology in human learning, we develop a novel ML framework which formalizes SGL into a machine learning skill. In our framework, each learner uses its intermediately trained model to generate a pseudo-labeled dataset and re-trains its model using  pseudo-labeled datasets generated by other learners. 
\item To formalize SGL, we develop a multi-level optimization (MLO) framework consisting of three learning stages: learners learn independently; learners learn collaboratively; learners validate themselves. 
\item An efficient optimization algorithm is developed to solve the MLO problem. 
\item We apply SGL for neural architecture search. The effectiveness of SGL is demonstrated by the experimental results on CIFAR-100, CIFAR-10, and ImageNet. 
\end{itemize}

\section{Related Works}
\paragraph{Neural Architecture Search (NAS).} NAS aims to  identify highly-performing architectures of deep neural networks automatically instead of manually designing them by humans. Various approaches have been proposed for NAS, including differentiable search methods~\citep{cai2018proxylessnas,liu2018darts,xie2018snas} and those based on reinforcement learning \citep{zoph2016neural,pham2018efficient,zoph2018learning} and evolutionary algorithms~\citep{liu2017hierarchical,real2019regularized}.  In RL-based approaches, a policy is learned to iteratively generate new architectures by maximizing a reward which is the accuracy on the validation set. Evolutionary algorithm based approaches represent  architectures as individuals in a population. Individuals with high fitness scores (validation accuracy) have the privilege to generate offspring, which replaces individuals with low fitness scores. Differentiable  approaches adopt a network pruning strategy. On top of an over-parameterized network,  importance weights of building blocks are learned using gradient descent. After learning, blocks whose weights are close to zero are  pruned. There have been many efforts devoted to improving differentiable NAS methods. In P-DARTS \citep{chen2019progressive}, the depth of searched architectures is allowed to grow progressively during the training process.   Search space approximation and regularization approaches are developed to reduce computational overheads and improve search stability.  PC-DARTS \citep{abs-1907-05737} reduces the redundancy in exploring the search space by sampling a small portion  of a super network. Operation search is performed in a subset of channels with the held-out part bypassed in a shortcut. Our proposed SGL framework is applicable to any differentiable search approach.  
\paragraph{Pseudo Labeling.} Pseudo labeling has been broadly applied to different applications including knowledge distillation~\citep{hinton2015distilling}, adversarial robustness~\citep{carlini2017towards}, self-supervised learning~\citep{xie2020self}, etc. \citet{zhang2018deep} proposed a deep mutual learning (DML) approach, which applies pseudo labeling for mutually learning multiple learners. Our proposed method differs from this one in two aspects. First, our method proposes a multi-level optimization framework to perform pretraining, pseudo-labeling, and finetuning (based on pseudo-labels) jointly end-to-end, where the models are pretrained before they are used for generating pseudo labels. In contrast, DML directly uses untrained models to perform pseudo-labeling, which may incur collective failures as discussed in Section \ref{sec:ablation}. Second, our method focuses on searching neural architectures while in DML the architectures are manually designed by humans.  In several works~\citep{GuT20,LiPYWLLC20,abs-2006-08341},  a trained teacher network (with a fixed architecture) is leveraged to generate pseudo-labels, which are used to search the architecture of a student network. \citet{abs-1912-07768} proposed a meta-learning approach to learn a generative model which generates synthetic data and uses the generated data to search neural architectures. In these works, pseudo-labeling is unidirectional: a network $A$ with a fixed architecture generates pseudo-labels which are used to search the architecture of $B$; $B$ does not generate pseudo-labels to search the architecture of $A$. Different from these works, in our work pseudo-labeling is bi-directional: each model generates pseudo-labels which are used to search the architectures of other models; meanwhile, each model uses pseudo-labels generated by other models to search its own architecture. 
\paragraph{Ensemble Learning.} Our work is also related to ensemble learning, such as boosting~\citep{freund1997decision}, bagging~\citep{breiman2001random}, etc. In ensemble learning, a set of models are learned to perform the same task. During training, these models do not necessarily collaborate with each other. During testing, the trained models work together to make a single prediction. The individual models in ensemble learning are preferred to be different from each other so that their comparative advantages can be combined together when making predictions. 
Our work differs from  ensemble learning in that the learners in our method help with each other during training; after training, different models achieve similar performance as a result of mutual teaching; during testing, only the single best model is retained for making predictions. Our method is more computationally efficient and memory efficient than ensemble learning methods during testing time since a single model is used for making predictions while  ensemble learning utilizes all models to perform predictions. 
\paragraph{Cooperative Learning.} Previously, cooperative learning, where multiple models collaborate to perform a task, has been investigated in multi-agent learning~\citep{panait2005cooperative}, co-training~\citep{blum1998combining}, federated learning~\citep{bonawitz2019towards}, etc. Our method differs from these existing methods in that we focus on collaborative neural architecture search via mutual pseudo-labeling.

\section{Methods}
In this section, we propose a small-group learning (SGL) framework and develop an optimization algorithm.

\begin{table}[t]
\caption{Notations in small-group learning}
\centering
\begin{tabular}{l|l}
\hline
Notation & Meaning \\
\hline
$A_k$ & Architecture of learner $k$\\
$V_k$ & The first set of  weights of learner $k$\\
$W_k$ & The second set of  weights of learner $k$\\
$D^{(\textrm{tr})}$ & Training dataset\\
$D^{(\textrm{val})}$ & Validation dataset\\
$D^{(\textrm{u})}$ & Unlabeled dataset \\
\hline
\end{tabular}
\label{tb:notations}
\end{table}

\subsection{Small-Group Learning}
In our framework, there are a set of $K$ learners, all of which learn to solve the same target task. Without loss of generality, we assume the task is classification. 
Each learner $k$ has a learnable architecture $A_k$ and two sets of learnable network weights $V_k$ and $W_k$. All learners share the same training dataset $D^{(\textrm{tr})}$, the same validation dataset $D^{(\textrm{val})}$, and an unlabeled dataset $D^{(\textrm{u})}$. 
The $K$ learners perform learning in three stages.  In the first stage, each learner trains a preliminary model. In the second stage, each learner uses its model trained in the first stage to perform pseudo-labeling. Then each learner re-trains its model using pseudo-labeled datasets generated by other learners. In the third stage, each learner measures the validation performance of its model trained in the second stage and updates its architecture to improve the validation performance. We discuss the details in the sequel. 
In the first stage, each learner $k$ trains its weights  $V_k$, with its architecture $A_k$ fixed:
\begin{equation}
    V^*_k(A_k) =\textrm{min}_{V_k} \; L(V_k,A_k,D^{(\textrm{tr})}). 
\end{equation}
The architecture $A_k$ is used to make predictions on training examples and define the training loss. However, $A_k$ should not be optimized to minimize the training loss. Otherwise, the training loss can be easily minimized to zero by making $A_k$ a very large (hence highly expressive) architecture, which will overfit the training data and have poor performance on unseen data. The optimally trained weights $V^*_k(A_k)$ is a function of $A_k$ since $V^*_k(A_k)$ is a function of $L(V_k,A_k,D^{(\textrm{tr})})$ and $L(V_k,A_k,D^{(\textrm{tr})})$ is a function of $A_k$.

\begin{figure}[t]
    \centering
 \includegraphics[width=\columnwidth]{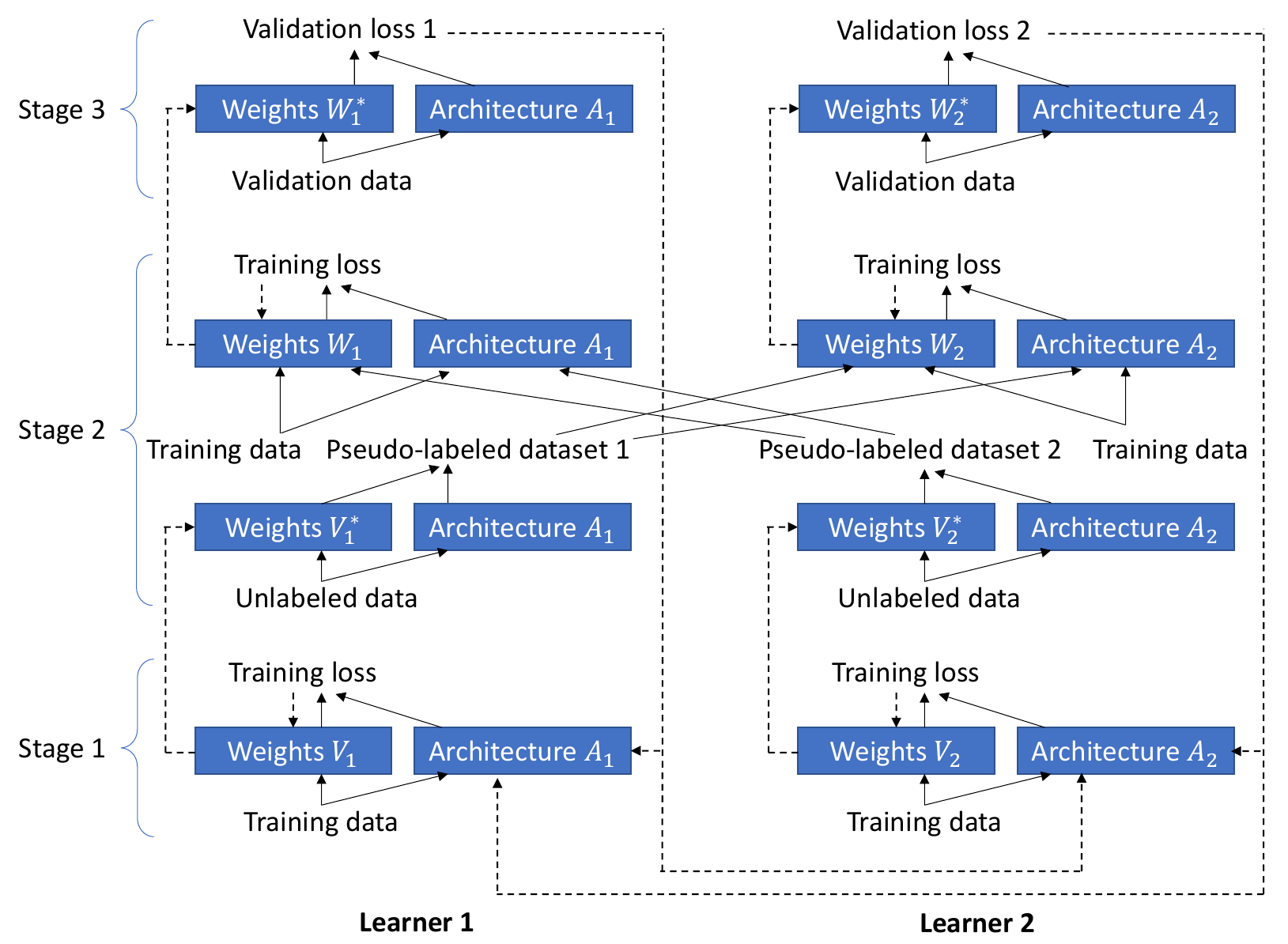}
       \caption{Three learning stages in small-group learning. Solid arrows denote the forward pass where predictions are made and losses are calculated.  Dotted arrows represent the backward pass of calculating gradients of model architectures and weights and performing gradient-based updates. 
        For simplicity, we assume there are two learners in the group. Extension to multiple learners is  straightforward.}
 \label{fig:arch}
\end{figure}

In the second stage, each learner $k$ uses  $V^*_k(A_k)$ learned in the first stage to make predictions on an  unlabeled dataset  $D^{(\textrm{u})}=\{x_i\}_{i=1}^N$ and generates a pseudo-labeled dataset $D^{(\textrm{pl})}_k(D^{(\textrm{u})},V^*_k(A_k))=\{(x_i,f(x_i;V^*_k(A_k)))\}_{i=1}^N$, where $f(\cdot;V^*_k(A_k))$ is the network parameterized by $V^*_k(A_k)$. $f(x_i;V^*_k(A_k))$ is a $J$-dimensional vector where $J$ is the number of classes and the $j$-th element of $f(x_i;V^*_k(A_k))$ indicates the probability that $x_i$ belongs to the $j$-th class. The sum of all elements in $f(x_i;V^*_k(A_k))$ is one. Meanwhile, each learner $k$ uses pseudo-labeled datasets $\{D^{(\textrm{pl})}_j\}_{j\neq k}^K$ produced by other learners as well as a human-labeled dataset $D^{(\textrm{tr})}$ to train its  other set of network weights $W_k$:
\begin{equation*}
\begin{array}{l}
    W^*_k(A_k,\{V^*_j(A_j)\}_{j\neq k}^K)
    =
    \underset{W_k}{\textrm{min}}
     \;   L(W_k, A_k, D^{(\textrm{tr})} )
     +\lambda \sum\limits_{j\neq k}^K L(W_k, A_k, D^{(\textrm{pl})}_j(D^{(\textrm{u})},V^*_j(A_j)) ),
     \end{array}
\end{equation*}
where the first loss term $L(W_k, A_k, D^{(\textrm{tr})} )$ in the objective is defined on the human-labeled training dataset and the second loss term is defined on the pseudo-labeled datasets produced by other learners. Both losses are cross-entropy losses. $\lambda$ is a tradeoff parameter.
Due to the same reason mentioned above, $A_k$ should not be updated at this stage either. 
Note that $W^*_k(A_k,\{V^*_j(A_j)\}_{j\neq k}^K)$ is a function of $A_k$ and $\{V^*_j(A_j)\}_{j\neq k}^K$ since
$W^*_k(A_k,\{V^*_j(A_j)\}_{j\neq k}^K)$ is a function of $L(W_k, A_k, D^{(\textrm{tr})} )
     +\lambda \sum_{j\neq k}^K L(W_k, A_k, D^{(\textrm{pl})}_j(D^{(\textrm{u})},V^*_j(A_j)) )$ which is a function of $A_k$ and $\{V^*_j(A_j)\}_{j\neq k}^K$.
In the third stage, each learner validates its   $W^*_k(A_k,\{V^*_j(A_j)\}_{j\neq k}^K)$ on the validation set $D^{(\textrm{val})}$. The learners optimize their architectures by minimizing the validation losses:
\begin{equation}
\underset{\{A_k\}_{k=1}^K}{\textrm{min}}
    \; \sum\limits_{k=1}^KL(W^*_k(A_k,\{V^*_j(A_j)\}_{j\neq k}^K), A_k,D^{(\textrm{val})}).
\end{equation}
Performed in a unified framework, the three stages mutually influence each other: $\{V^*_k(A_k)\}_{k=1}^K$ trained in the first stage are leveraged to calculate the training loss in the second stage; $\{W^*_k(A_k,\{V^*_j(A_j)\}_{j\neq k}^K)\}_{k=1}^K$ trained in the second stage are leveraged to calculate the validation loss in the third stage; after $\{A_k\}_{k=1}^K$ are optimized  in the third stage, they will render  the training loss in the first  stage to be changed, which changes $\{V^*_k(A_k)\}_{k=1}^K$ in the second stage accordingly. Figure~\ref{fig:arch} illustrates the three stages.
To this end, we formulate SGL as a three-level optimization problem:
\begin{equation}
\begin{array}{l}
\underset{\{A_k\}_{k=1}^K}{\textrm{min}}
    \; \sum\limits_{k=1}^KL(W^*_k(A_k,\{V^*_j(A_j)\}_{j\neq k}^K),A_k, D^{(\textrm{val})})\\
      s.t. \;\;\;    \{W^*_k(A_k,\{V^*_j(A_j)\}_{j\neq k}^K)\}_{k=1}^K 
    =\\
    \quad\quad
    \underset{\{W_k\}_{k=1}^K}{\textrm{min}}
     \;  \sum\limits_{k=1}^K L(W_k, A_k, D^{(\textrm{tr})} )   +\lambda \sum\limits_{j\neq k}^K L(W_k, A_k, D^{(\textrm{pl})}_j(D^{(\textrm{u})},V^*_j(A_j)) )\\
    \quad\quad \{V^*_k(A_k)\}_{k=1}^K  =
    \underset{\{V_k\}_{k=1}^K}{\textrm{min}} \; \sum\limits_{k=1}^K L(V_k,A_k,D^{(\textrm{tr})})
\end{array}
\label{eq:sgl}
\end{equation}
This formulation consists of three optimization problems including two inner-optimization problems and one outer-optimization problem.  The two inner-optimization problems are on the constraints of  the outer optimization problem. From bottom to top, the three optimization problems correspond to the first, second, and third learning stage respectively. Inspired by \citep{liu2018darts}, we parameterize the architecture $A$ using continuous variables which allow the search to be conducted using efficient gradient-based methods. The search space of $A$ is  overparameterized with many building blocks. For each block, a continuous variable is used to represent whether this block should be selected to form the final architecture. Search amounts to learning these variables.

\subsection{Optimization Algorithm}
\label{sec:alg}
To solve the SGL problem, we develop an efficient gradient-based optimization algorithm, drawing inspirations from~\citep{liu2018darts}. 
In the following, $\nabla^2_{Y,X}f(X,Y)$ denotes $\frac{\partial f(X,Y)}{\partial X\partial Y}$.  
First of all, we approximate $V_k^{*}(A_k)$ using 
\begin{equation}
    V'_k=V_k - \xi_{V}  \nabla_{V_k}L(V_k, A_k, D^{(\mathrm{tr})}),
    \label{eq:update_v}
\end{equation}
where $\xi_{V}$ is a learning rate. Plugging $\{V'_j\}_{j=1}^K$ into $\sum_{k=1}^K (L(W_k, A_k, D^{(\textrm{tr})} )+\lambda \sum_{j\neq k}^K L(W_k, A_k,$\\$ D^{(\textrm{pl})}_j(D^{(\textrm{u})},V^*_j(A_j)) ))$, we obtain an approximated objective $\sum_{k=1}^K O^{W}_k$ where $O^{W}_k=L(W_k, $\\$A_k, D^{(\textrm{tr})} )+\lambda \sum_{j\neq k}^K L(W_k, A_k, D^{(\textrm{pl})}_j(D^{(\textrm{u})},V'_j) )$. Then  we approximate  $W^*_k(A_k,\{V^*_j(A_j)\}_{j\neq k}^K)$ using one-step gradient descent update of  $W_k$ w.r.t $O^{W}_k$:
\begin{equation}
\begin{array}{l}
    W'_k=W_k - \xi_{W}  \nabla_{W_k}(L(W_k, A_k, D^{(\textrm{tr})} )+
  \lambda \sum_{j\neq k}^K L(W_k, A_k, D^{(\textrm{pl})}_j(D^{(\textrm{u})},V'_j) )).
    \label{eq:update_w}
    \end{array}
\end{equation}
Finally, we plug $\{W'_k\}_{k=1}^K$ into $\sum_{k=1}^KL(W^*_k(A_k,\{V^*_j(A_j)\}_{j\neq k}^K),A_k, D^{(\textrm{val})})$ and get $\sum_{k=1}^KL(W'_k,$\\$A_k, D^{(\textrm{val})})$. We can update the architecture $A_k$ of learner $k$ by descending the gradient of  $\sum_{j=1}^KL(W'_j,A_j, D^{(\textrm{val})})$ w.r.t $A_k$:
\begin{equation}
\begin{array}{l}
A_k\gets A_k-\eta_{A} ( \nabla_{A_k} L(W'_k,A_k, D^{(\textrm{val})})+\sum\limits_{j\neq k}^K \nabla_{A_k} L(W'_j,A_j, D^{(\textrm{val})})).
    \end{array}
    \label{eq:update_a}
\end{equation}
where 
\begin{equation}
\begin{array}{l}
    \nabla_{A_k} L(W'_k,A_k, D^{(\textrm{val})})=\\
      \nabla_{A_k} L(W_k - \xi_{W}  \nabla_{W_k}(L(W_k, A_k, D^{(\textrm{tr})} )+\lambda \sum_{j\neq k}^K L(W_k, A_k, D^{(\textrm{pl})}_j(D^{(\textrm{u})},V'_j) )),A_k, D^{(\textrm{val})})=\\
 \nabla_{A_k}L(W'_k,A_k, D^{(\textrm{val})}) - \xi_{W}  \nabla^2_{A_k,W_k}(L(W_k, A_k, D^{(\textrm{tr})} )+\\\lambda \sum_{j\neq k}^K L(W_k, A_k, D^{(\textrm{pl})}_j(D^{(\textrm{u})},V'_j) ) )  \nabla_{W'_k}L(W'_k,A_k, D^{(\textrm{val})}). \\
    \end{array}
\end{equation}
Calculating 
 matrix-vector product $\nabla^2_{A_k,W_k}(L(W_k, A_k, D^{(\textrm{tr})} )+\lambda \sum_{j\neq k}^K L(W_k, A_k, D^{(\textrm{pl})}_j(D^{(\textrm{u})}$\\$,V'_j) ))\nabla_{W'_k}L(W'_k,A_k, D^{(\textrm{val})})$ incurs a lot of computational cost. This cost can be reduced  by a finite difference approximation~\citep{liu2018darts}:
\begin{equation}
\begin{array}{ll}
     \nabla^2_{A_k,W_k}(L(W_k, A_k, D^{(\textrm{tr})} )+\lambda \sum_{j\neq k}^K L(W_k, A_k, D^{(\textrm{pl})}_j(D^{(\textrm{u})},V'_j) ))   \nabla_{W'_k}L(W'_k,A_k, D^{(\textrm{val})})\approx  \\
     \frac{1}{2\alpha}(\nabla_{A_k} (L(W^+_k, A_k, D^{(\textrm{tr})} )+\lambda \sum\limits_{j\neq k}^K L(W^+_k, A_k, D^{(\textrm{pl})}_j(D^{(\textrm{u})},V'_j) ))\\
     -\nabla_{A_k} (L(W^-_k, A_k, D^{(\textrm{tr})} )+\lambda \sum\limits_{j\neq k}^K L(W^-_k, A_k, D^{(\textrm{pl})}_j(D^{(\textrm{u})},V'_j) ))),
\end{array}
\label{eq:finite-aw}
\end{equation}
where $W^{\pm}_k=W_k \pm \alpha \nabla_{W^{\prime}_k} L(W'_k,A_k, D^{(\textrm{val})})$ and $\alpha$ is a small scalar $0.01 /\|\nabla_{W'_k} L(W'_k,A_k,D^{(\textrm{val})})\|_{2}$.
For $\nabla_{A_k} L(W'_j,A_j, D^{(\textrm{val})}) (j\neq k)$ in Eq.(\ref{eq:update_a}), it can be calculated as:
\begin{equation}
\begin{array}{ll}
    \nabla_{A_k} L(W'_j,A_j, D^{(\textrm{val})})=\\
   \frac{\partial W'_j}{\partial A_k} \nabla_{W'_j} L(W'_j,A_j, D^{(\textrm{val})})=\\
   - \xi_{W}\lambda
   
   \frac{\partial \nabla_{W_j} L(W_j, A_j, D^{(\textrm{pl})}_k(D^{(\textrm{u})},V'_k) )}{\partial A_k}

   \nabla_{W'_j} L(W'_j,A_j, D^{(\textrm{val})})=\\
   
    - \xi_{W}\lambda

   \frac{\partial V'_k }{\partial A_k}

 \nabla^2_{V'_k,W_j} L(W_j, A_j, D^{(\textrm{pl})}_k(D^{(\textrm{u})},V'_k) )
   
   \nabla_{W'_j} L(W'_j,A_j, D^{(\textrm{val})})\\
   
    \end{array}
\end{equation}
where 
\begin{align}
    \frac{\partial V'_k}{\partial A_k}&=\frac{\partial (V_k - \xi_{V}  \nabla_{V_k}L(V_k, A_k, D^{(\mathrm{tr})}))}{\partial A_k}\\
    &= -\xi_{V}\nabla^2_{A_k,V_k}L(V_k, A_k, D^{(\mathrm{tr})})
\end{align}

The algorithm for solving SGL is  in Algorithm~\ref{algo:algo}.
\begin{algorithm}[H]
\SetAlgoLined
 \While{not converged}{
1. For learner $1,\cdots,K$, update  $V_k$ using Eq.(\ref{eq:update_v})\\
2. For learner $1,\cdots,K$, update  $W_k$ using Eq.(\ref{eq:update_w})\\
3. For learner $1,\cdots,K$, update $A_k$ using Eq.(\ref{eq:update_a})
 }
 \caption{Optimization algorithm for SGL}
 \label{algo:algo}
\end{algorithm}

\section{Experiments}
In this section, our proposed SGL is applied to search neural architectures for image classification.  Please refer to the supplements for more hyperparameter settings, results, and significance tests of results.

\subsection{Datasets}
We performed the experiments on three datasets: CIFAR-10, CIFAR-100, and ImageNet~\citep{deng2009imagenet}. CIFAR-10 is split into a 25K training set, a 25K validation set, and a 10K test set. So is CIFAR-100. The training and validation set is used as $D^{\textrm{(tr)}}$ and $D^{\textrm{(val)}}$ in SGL. 
ImageNet has 1.3M training images and 50K test images. CIFAR-10 and CIFAR-100 have 10 classes and ImageNet has 1000 classes.

\subsection{Experimental Settings}
Following the protocol in \citep{liu2018darts}, each experiment consists of an architecture search phrase and an architecture evaluation phrase. In the search phrase, an architecture $A$ is learned by solving Eq.(\ref{eq:sgl}). In the evaluation phrase, multiple copies of $A$ are composed into a larger network, which is then trained from scratch and tested on the test set. 
For the search space of $A$, we used the ones proposed in DARTS~\citep{liu2018darts}, 2) DARTS$^{-}$ \citep{abs-2009-01027}, 3) P-DARTS~\citep{chen2019progressive}, and 4) PC-DARTS~\citep{abs-1907-05737}, where the building blocks include  $3\times 3$ and $5\times 5$ (dilated) separable convolutions, $3\times 3$ max/average pooling, identity operation, and zero operation. In SGL, we set the number of learners to 2. The tradeoff parameter $\lambda$ was tuned in $\{0.1,0.5,1,2,3\}$ on a held-out 5k dataset. The best performing value of  $\lambda$ is 1. When searching architectures on CIFAR-10, the input images (without labels) in CIFAR-100 were used as the unlabeled dataset; vice versa.

For architecture search on CIFAR-10 and CIFAR-100, each architecture consists of a stack of 8 cells and each cell consists of 7 nodes. The initial channel number was set to 16. The rest hyperparameters for  architectures and network weights mostly follow those in DARTS, P-DARTS, PC-DARTS, and DARTS$^{-}$. 
The Adam optimizer~\citep{adam} was used to optimize architecture variables, with a learning rate of 3e-4, a batch size of 50, and a weight decay of 1e-3. The SGD optimizer was used to optimize network weights, with an initial learning rate of 0.025, a cosine decay scheduler, a batch size of 50, a weight decay of 3e-4, and  a momentum of 0.9. The search algorithm ran for 50 epochs. 
For architecture search on ImageNet, following~\citep{abs-1907-05737}, we randomly sample 10\% images from the 1.3M training set as $D^{\textrm{(tr)}}$ and 2.5\% images as $D^{\textrm{(val)}}$ in SGL. Another randomly sampled 10\% images (excluding labels) are used as the unlabeled dataset. The hyperparameters in the ImageNet experiments mostly follow those in the CIFAR-10 experiments. These experiments were conducted using a single Tesla v100 GPU.

After searching, among the $K$ learners, the one achieving the smallest validation loss is retained and the rest are discarded. The architecture $A^*$ of the retained learner is evaluated. 
For CIFAR-10 and CIFAR-100, the optimal cell in $A^*$ is duplicated 20 times, which are then stacked into a large network. The initial channel number was set to 36. 
This large network was trained on the combination of training set and validation set for 600 epochs. The SGD optimizer is used for weights training, with an initial learning rate of 0.025, a cosine decay scheduler, a batch size of 96, a momentum of 0.9, and a weight decay of 3e-4. 
For architecture evaluation on ImageNet, the large network is a stack of 14 cells, with an initial channel number of 48. The large network was trained on the 1.3M training images using SGD on 8 Tesla v100 GPUs. 
for 250 epochs, with an initial learning rate of 0.5, a batch size of 1024, and a weight decay of 3e-5.  
On ImageNet, we also evaluated the architectures searched on CIFAR-10 and CIFAR-100, following the same evaluation protocol. We repeat each SGL experiment 10 times with different random seeds (between 1 and 10). The mean and standard deviation of test results in the 10 runs are reported.

\subsection{Results}

\begin{table}[t]
\caption{Test error, parameter number, and search cost (GPU days) on CIFAR-100. Search cost is measured by GPU days on a Tesla v100. In SGL, we count the number of parameters in the single retained learner (the one yielding the lowest validation loss).
    DARTS(1st) and DARTS(2nd) denotes that the first-order and second-order approximation is used in DARTS. On CIFAR-100, DARTS(2nd) is no better than DARTS(1st) despite using a much more computationally-heavy and memory-inefficient second-order approximation. 
    * indicates that the results are taken from DARTS$^{-}$ \citep{abs-2009-01027}. $\dag$ indicates that the results were obtained by re-running the methods for 10 times. In our run of DARTS$^{-}$, we were not able to achieve the error  reported in~\citep{abs-2009-01027}. $\Delta$ indicates that during architecture evaluation, we ran for 600 epochs instead of 2000 as in \citep{liang2019darts+} to make sure the comparison with other approaches (which use 600 epochs) is fair. 
    }
    \centering
    \begin{tabular}{l|ccc}
    \toprule
    Method & Error(\%)& Param(M)& Cost\\
    \midrule
    *ResNet \citep{he2016deep}&22.10&1.7&-\\
     *DenseNet \citep{HuangLMW17}&17.18&25.6 &-\\
    \hline
    *PNAS \citep{LiuZNSHLFYHM18}&19.53&3.2&150\\
    *ENAS \citep{pham2018efficient}&19.43&4.6&0.5\\
        *AmoebaNet \citep{real2019regularized}&18.93&3.1&3150\\
    \hline
      *DARTS-2nd \citep{liu2018darts}  & 20.58$\pm$0.44&1.8&1.5 \\
    *GDAS \citep{DongY19}&18.38&3.4&0.2\\
    *R-DARTS \citep{ZelaESMBH20}&18.01$\pm$0.26&-&1.6
    \\
    ${}^{\Delta}$DARTS$^{+}$ \citep{abs-1909-06035}&17.11$\pm$0.43&3.8&0.2\\
      *DropNAS \citep{HongL0TWL020} & 16.39&4.4&0.7 \\
\hline
\hline
     ${}^{\dag}$DARTS(1st) \citep{liu2018darts}  &20.52$\pm$0.31 &1.8 &0.4\\
     $\;\;$DARTS(1st) + SGL (ours) &\textbf{18.54}$\pm$0.21 & 2.2&1.2 \\
            \hline
      *DARTS$^{-}$ \citep{abs-2009-01027}&17.51$\pm$0.25&3.3&0.4\\
      ${}^{\dag}$DARTS$^{-}$ \citep{abs-2009-01027}& 18.97$\pm$0.16& 3.1&0.4\\
           $\;\;$DARTS$^{-}$ + SGL (ours) & \textbf{16.90}$\pm$0.10&3.5&2.0\\
                   \hline
           *P-DARTS \citep{chen2019progressive}&17.49&3.6&0.3\\
       $\;\;$P-DARTS + SGL (ours) &  \textbf{16.58}$\pm$0.18&3.6 &2.1\\
     \hline
    ${}^{\dag}$PC-DARTS \citep{abs-1907-05737} & 17.01$\pm$0.06& 4.0& 0.1\\
     $\;\;$PC-DARTS + SGL (ours) &\textbf{16.34}$\pm$0.11&4.1&0.5   \\
        \bottomrule
    \end{tabular}
    \label{tab:cifar100}
\end{table}

\begin{table}[t]
  \caption{
    Test error, parameter number, and search cost (GPU days) on CIFAR-10.
    * indicates the results are taken from DARTS$^{-}$ \citep{abs-2009-01027}, NoisyDARTS~\citep{abs-2005-03566},  DrNAS~\citep{abs-2006-10355}, and GTN~\citep{abs-1912-07768}.
    Other notations are the same as those in Table~\ref{tab:cifar100}.
    }
    \centering
    \begin{tabular}{l|ccc}
    \toprule
    Method& Error(\%)& Param(M) & Cost\\
    \midrule
    *DenseNet
    \citep{HuangLMW17}&3.46&25.6 &-\\
    \hline
     *HierEvol \citep{liu2017hierarchical}&3.75$\pm$0.12& 15.7 &300\\
    *NAONet-WS \citep{LuoTQCL18} & 3.53 & 3.1&0.4 \\
        *PNAS \citep{LiuZNSHLFYHM18} &3.41$\pm$0.09  &3.2& 225\\
        *ENAS \citep{pham2018efficient} &2.89 & 4.6  &0.5 \\
    *NASNet-A \citep{zoph2018learning} & 2.65 & 3.3& 1800\\
    *AmoebaNet-B \citep{real2019regularized} & 2.55$\pm$0.05 & 2.8&3150  \\
    \hline
        *R-DARTS \citep{ZelaESMBH20} &2.95$\pm$0.21  &- & 1.6 \\
            *GDAS \citep{DongY19}&2.93& 3.4& 0.2 \\
    *SNAS \citep{xie2018snas} &2.85 & 2.8& 1.5\\
             \hline ${}^{\Delta}$DARTS$^{+}$ \citep{abs-1909-06035}&2.83$\pm$0.05&3.7&0.4\\
        *BayesNAS \citep{ZhouYWP19} &2.81$\pm$0.04 &3.4&0.2 \\
            *DARTS-2nd \citep{liu2018darts} &2.76$\pm$0.09&3.3&  1.5\\
        *MergeNAS \citep{WangXYYHS20} &2.73$\pm$0.02 &2.9 & 0.2 \\
        *NoisyDARTS \citep{abs-2005-03566} &2.70$\pm$0.23&3.3  & 0.4 \\
            *ASAP \citep{NoyNRZDFGZ20} &2.68$\pm$0.11 & 2.5&0.2 \\
                *SDARTS
    \citep{abs-2002-05283}&2.61$\pm$0.02 & 3.3& 1.3 \\
            *DropNAS \citep{HongL0TWL020} &2.58$\pm$0.14 & 4.1&0.6 \\
    *FairDARTS \citep{abs-1911-12126} &2.54 &3.3 &0.4 \\
       *DrNAS \citep{abs-2006-10355} &2.54$\pm$0.03&4.0&  0.4\\
    *GTN~\citep{abs-1912-07768}& 2.42$\pm$0.03 & 97.9&  0.67\\
    \hline
        \hline
            *DARTS(1st) \citep{liu2018darts} &3.00$\pm$0.14&3.3&  0.4\\
        $\;\;$DARTS(1st) + SGL (ours) & \textbf{2.41}$\pm$0.06&3.7&1.2 \\
            \hline
             *DARTS$^{-}$ \citep{abs-2009-01027}&2.59$\pm$0.08&  3.5&0.4\\
             ${}^{\dag}$DARTS$^{-}$ \citep{abs-2009-01027}& 2.97$\pm$0.04& 3.3&0.6\\
         $\;\;$DARTS$^{-}$ + SGL (ours) & 2.60$\pm$0.07&3.1&2.0 \\
             \hline
         *PC-DARTS \citep{abs-1907-05737} &2.57$\pm$0.07&3.6& 0.1\\
         $\;\;$PC-DARTS + SGL (ours) &2.60$\pm$0.12&3.5&0.5 \\
     \hline
    *P-DARTS \citep{chen2019progressive}& 2.50&3.4&  0.3\\
     $\;\;$P-DARTS + SGL (ours)& 2.47$\pm$0.10&3.6 & 2.1 \\
        \bottomrule
    \end{tabular}
    \label{tab:cifar10}
\end{table}

\begin{table}[t]
\caption{Top-1 and top-5 test errors (\%) on ImageNet.  * indicates that the results are taken from DARTS$^{-}$ \citep{abs-2009-01027}, DrNAS \citep{abs-2006-10355}, and AKDNet~\citep{LiuJTVZGW20}. $\dag$ denotes that the result is obtained from our run. The rest notations are the same as those in Table~\ref{tab:cifar100}. From top to bottom, different panels show manually-designed networks, non-differentiable search methods, and differentiable search methods. 
    }
    \centering
    \begin{adjustbox}{width=0.98\columnwidth,center}
     \begin{tabular}{l|cccc}
    \toprule
  \multirow{ 2}{*}{Method}   & Top-1  &Top-5 &Param & Runtime \\
         & Error (\%) & Error (\%)&(M) & (GPU days)\\
    \midrule
    *Inception-v1 \citep{googlenet}&30.2 &10.1&6.6&- \\
    *MobileNet \citep{HowardZCKWWAA17} &  29.4& 10.5 &4.2&- \\
    *ShuffleNet 2$\times$ (v1) \citep{ZhangZLS18} &  26.4 &10.2 & 5.4&-\\
    *ShuffleNet 2$\times$ (v2) \citep{MaZZS18} &  25.1 &7.6 & 7.4&-\\
    \hline
    *NASNet-A \citep{zoph2018learning} &26.0 &8.4 &5.3 &1800 \\
    *PNAS \citep{LiuZNSHLFYHM18} &25.8 &8.1  &5.1 &225 \\
    *MnasNet-92 \citep{TanCPVSHL19} & 25.2 & 8.0& 4.4&1667\\
        *AmoebaNet-C \citep{real2019regularized} &  24.3 &7.6 &6.4&3150 \\
    \hline
     *SNAS-CIFAR10 \citep{xie2018snas} & 27.3 &9.2 &4.3 &1.5 \\
          *BayesNAS-CIFAR10 \citep{ZhouYWP19} &26.5 &8.9 &3.9&0.2 \\
                    *PARSEC-CIFAR10 \citep{abs-1902-05116} & 26.0 &8.4&5.6&1.0 \\
     *GDAS-CIFAR10 \citep{DongY19} &  26.0&8.5 &5.3 & 0.2\\
                 *DSNAS-ImageNet \citep{HuXZLSLL20} &25.7& 8.1 &- & -\\
          *SDARTS-ADV-CIFAR10 \citep{abs-2002-05283}&25.2& 7.8 &5.4& 1.3 \\
           *PC-DARTS-CIFAR10 \citep{abs-1907-05737} & 25.1 &7.8&5.3&0.1\\
                *ProxylessNAS-ImageNet \citep{cai2018proxylessnas} & 24.9 &7.5 &7.1 &8.3  \\
          *FairDARTS-CIFAR10 \citep{abs-1911-12126} &24.9 &7.5 &4.8 &0.4 \\
     *FairDARTS-ImageNet \citep{abs-1911-12126} &24.4 &7.4 &4.3 &3.0 \\
             *DrNAS-ImageNet \citep{abs-2006-10355} & 24.2 &7.3& 5.2&3.9\\
         *DARTS$^{+}$-ImageNet \citep{abs-1909-06035}& 23.9& 7.4&5.1&6.8\\
        *DARTS$^{-}$-ImageNet \citep{abs-2009-01027}&23.8& 7.0&4.9&4.5\\
     *DARTS$^{+}$-CIFAR100 \citep{abs-1909-06035}&23.7& 7.2&5.1&0.2\\
     \hline
       \hline
            *DARTS-2nd-CIFAR10 \citep{liu2018darts}  & 26.7 &8.7&4.7&0.4 \\ $\dag$DARTS-1st-CIFAR10 \citep{liu2018darts}  & 26.1 &8.3 &4.5 &0.4  \\
        $\;\;$SGL-DARTS-1st-CIFAR10 (ours) & \textbf{24.9} & \textbf{7.7}&5.2 &2.1  \\
        \hline
          *P-DARTS-CIFAR10 \citep{chen2019progressive}&24.4 &7.4&4.9&0.3\\
        $\;\;$SGL-P-DARTS-CIFAR10 (ours) &\textbf{24.3}  &\textbf{7.2}  &5.1 &2.1  \\
        \hline
             *P-DARTS-CIFAR100 \citep{chen2019progressive}&24.7& 7.5&5.1&0.3\\
           $\;\;$SGL-P-DARTS-CIFAR100 (ours) &\textbf{23.9}  & \textbf{7.2} & 5.3&2.1 \\
           \hline
            *PC-DARTS-ImageNet \citep{abs-1907-05737} &  24.2 &7.3&5.3&3.8\\
          $\;\;$SGL-PC-DARTS-ImageNet (ours)& \textbf{23.3} & \textbf{6.7} & 6.4&4.0 \\
        \bottomrule
    \end{tabular}
    \end{adjustbox}
    \label{tab:imagenet}
\end{table}

In Table~\ref{tab:cifar100}, we compare the classification error on the test set, number of weight parameters, and search cost (GPU days) of different methods on CIFAR-100. We observe the following from this table. \textbf{First}, under different settings of search spaces, including those from DARTS, DARTS$^{-}$,  P-DARTS, and PC-DARTS, our method achieves significantly lower test errors than the original baselines. For example, DARTS(1st)+SGL achieves an error of 18.54\% which is greatly lower than the  20.52\% error of DARTS(1st). As another example, PDARTS+SGL achieves a significantly lower error of 16.58\%, compared with the 17.49\% error of P-DARTS. These results show that our proposed small-group learning is an effective approach in searching better-performing architectures.  In our method, learners with different architectures collaboratively solve the same task. Since having different architectures, these learners possess complementary advantages: for a certain class $x$, some learners perform better in classifying examples belonging to $x$ than other learners; for another class $y$, another set of learners have better classification abilities than the rest of learners; and so on. Via pseudo-labeling, each learner can transfer the knowledge in areas it is good at to other learners; equivalently, each learner can mitigate its deficiency in certain areas by leveraging the wisdom from other learners which are excellent in those areas. This collaboration mechanism enables different learners to jointly improve. In single-learner NAS, such a mechanism is lacking, which renders the performance to be inferior. \textbf{Second}, our method is broadly applicable and effective, as evidenced by the fact that our method improves a variety of baselines (including DARTS,  DARTS$^{-}$,  P-DARTS, and PC-DARTS) when applied to these methods. Our method is designed as a general framework which is agnostic to how the search space is crafted, which paves a way for leveraging our method to improve most (if not all)  existing differentiable NAS methods. \textbf{Third}, among all methods, PCDARTS+SGL achieves the best performance. This shows that our proposed SGL approach is very competitive in pushing the limit of NAS research.  \textbf{Fourth},  parameter number and search cost of our method is similar to those of other differentiable  methods. This indicates that our method can search more accurate architectures without incurring significant additional costs such as memory footprint, training time, inference time, etc.

In Table~\ref{tab:cifar10}, we compare different methods on CIFAR-10, in terms of classification error on test set, number of network weights, and search cost (GPU days). When applied to  DARTS(1st), our method improves the performance of this baseline very significantly, reducing the 3.00\% error of DARTS(1st) to 2.41\%, which surpasses all other methods in this table. This again demonstrates that our method is effective in searching highly-performant architectures, thanks to its mechanism of enabling models with complementary advantages to collaboratively learn from each other and mitigate weakness by taking in knowledge from stronger learners. The performance of GTN is close to ours. However, this method leads to a very large architecture with 97.9M parameters, which is about 26 times larger than ours. 
 When applied to other methods including ${}^{\dag}$DARTS$^{-}$, PC-DARTS, and P-DARTS, our method does not achieve a significant improvement, which is probably due to the fact that CIFAR-10 is a much easier classification dataset than CIFAR-100 and therefore different methods tend to achieve very close performance.

In Table~\ref{tab:imagenet}, we make a comparison of different methods on ImageNet, in terms of top-1 and top-5 classification errors. In SGL-PC-DARTS-ImageNet, our proposed SGL is applied to  PC-DARTS and the search is performed on ImageNet. Our method achieves a top-1 error of 23.3\% and a top-5 error of 6.7\%, which are the lowest among all methods in this table and  are much lower than those of PC-DARTS-ImageNet. This  demonstrates that our method has strong capability in searching high-quality architectures not only for small datasets like CIFAR10/100, but also for large-scale real-world datasets such as ImageNet. In addition, on ImageNet we evaluated the architectures searched on CIFAR10/100, including SGL-DARTS-1st-CIFAR10, SGL-P-DARTS-CIFAR10, and SGL-P-DARTS-CIFAR100. As can be seen, these architectures searched by our SGL method achieve better performance than those searched by corresponding baselines. For example, applying SGL to DARTS-1st-CIFAR10, we reduce the 26.1\% error of DARTS-1st-CIFAR10 down to 24.9\%. This further demonstrates the effectiveness of our method.

\subsection{Ablation Studies}
\label{sec:ablation}
In this section, we perform ablation studies to evaluate the importance of individual components in our SGL framework, by comparing the full SGL method with the following ablation settings. 
\begin{itemize}[leftmargin=*]
    \item \textbf{Ablation setting 1}. In this setting, the first learning stage in Eq.(\ref{eq:sgl}) is removed. Pseudo-labeling is performed  using the weights $W$. The corresponding formulation is:
    \begin{equation}
\begin{array}{l}

\underset{\{A_k\}_{k=1}^K}{\textrm{min}}
    \; \sum\limits_{k=1}^KL(W^*_k(\{A_j\}_{j=1 }^K),A_k, D^{(\textrm{val})})\\
      s.t. \;\;\;   \quad \{W^*_k(\{A_j\}_{j=1 }^K)\}_{k=1}^K 
    =
    \underset{\{W_k\}_{k=1}^K}{\textrm{min}}
     \;  \sum\limits_{k=1}^K L(W_k, A_k, D^{(\textrm{tr})} ) \\
    \qquad\quad +\lambda \sum\limits_{j\neq k}^K L(W_k, A_k, D^{(\textrm{pl})}_j(D^{(\textrm{u})},W_j, A_j) )\\
   \end{array}
\label{eq:ab-1}
\end{equation}
     In this experiment, $\lambda$ is set to 2 for CIFAR-10 and to 1 for CIFAR-100. In terms of the search space, it is set to the space used in P-DARTS for CIFAR-100, and set to the space used in DARTS(1st) for CIFAR-10. 
\item \textbf{Ablation setting 2}. In this setting, we separate the first stage in SGL from the second stage. We first run the first stage to search an architecture $B_k$ for each learner $k$ independently, by solving the following problem:
\begin{equation}
\begin{array}{l}
\underset{\{B_k\}_{k=1}^K}{\textrm{min}}
    \; \sum\limits_{k=1}^KL(V^*_k(B_k),B_k, D^{(\textrm{val})})\\
      s.t. \quad\;\;    \{V^*_k(B_k)\}_{k=1}^K  =
    \underset{\{V_k\}_{k=1}^K}{\textrm{min}} \; \sum\limits_{k=1}^K L(V_k,B_k,D^{(\textrm{tr})})
\end{array}
\label{eq:ab-21}
\end{equation}
Then we use the searched architectures $\{B^*_k\}_{k=1}^K$ together with their optimally trained network weights $\{V^*_k\}_{k=1}^K$ to perform pseudo-labeling and use the pseudo-labeled datasets $\{D^{(\textrm{pl})}_j(D^{(\textrm{u})},V^*_j,B^*_j)\}_{j=1}^K$ to perform the second stage, which amounts to solving the following problem:
\begin{equation}
\begin{array}{l}
\underset{\{A_k\}_{k=1}^K}{\textrm{min}}
    \; \sum\limits_{k=1}^KL(W^*_k(A_k),A_k, D^{(\textrm{val})})\\
      s.t. \quad\;\;    \{W^*_k(A_k)\}_{k=1}^K 
    =
    \underset{\{W_k\}_{k=1}^K}{\textrm{min}}
     \;  \sum\limits_{k=1}^K L(W_k, A_k, D^{(\textrm{tr})} )  +\lambda \sum\limits_{j\neq k}^K L(W_k, A_k, D^{(\textrm{pl})}_j(D^{(\textrm{u})},V^*_j,B^*_j) )
\end{array}
\label{eq:sgl}
\end{equation}
 In this experiment, $\lambda$ is set to 2 for CIFAR-10 and to 1 for CIFAR-100. In terms of the search space, it is set to the space used in P-DARTS for CIFAR-100, and set to the space used in DARTS(1st) for CIFAR-10.
  \item \textbf{Ablation setting 3}. In this setting, in the second stage of SGL, each learner is trained solely based on pseudo-labeled datasets by other learners, without using human-labeled  training data. 
  The corresponding formulation is: 
\begin{equation}
\begin{array}{l}
\underset{\{A_k\}_{k=1}^K}{\textrm{min}}
    \; \sum\limits_{k=1}^KL(W^*_k(A_k,\{V^*_j(A_j)\}_{j\neq k}^K),A_k, D^{(\textrm{val})})\\
      s.t. \;\;\;    \{W^*_k(A_k,\{V^*_j(A_j)\}_{j\neq k}^K)\}_{k=1}^K 
    =\underset{\{W_k\}_{k=1}^K}{\textrm{min}}
     \;  \sum\limits_{k=1}^K  \sum\limits_{j\neq k}^K L(W_k, A_k, D^{(\textrm{pl})}_j(D^{(\textrm{u})},V^*_j(A_j)) )\\
    \quad\quad \{V^*_k(A_k)\}_{k=1}^K  =
    \underset{\{V_k\}_{k=1}^K}{\textrm{min}} \; \sum\limits_{k=1}^K L(V_k,A_k,D^{(\textrm{tr})})
\end{array}
\label{eq:ab-2}
\end{equation}
For CIFAR-100, SGL is applied to P-DARTS. For CIFAR-10, SGL is applied to DARTS-1st.
    \item Ablation study on the tradeoff parameter $\lambda$.
    We investigate how the classification error changes with $\lambda$ in Eq.(\ref{eq:sgl}). 
    For either CIFAR-10 or CIFAR-100, 5K examples are uniformly sampled from the 25K training set and  25K validation set. Performance is reported on the 5K sampled data. The rest data is used as before. 
    SGL is applied to P-DARTS.  
    \item Ablation study on the number of learners $K$. We investigate how the classification error changes with the number of learners $K$. Performance is reported on the 5K randomly sampled data. SGL is applied to DARTS(1st). 
\end{itemize}

Table~\ref{tab:ab1} shows the classification errors on the test sets of CIFAR-10 and CIFAR-100, under the settings of ``with first stage" and ``without first stage" in the first ablation study. As can be seen, removing the  first stage renders  the errors to increase on both CIFAR-10 and CIFAR-100, compared with using the first stage. The reason is that: for ``without first stage", different learners directly use  untrained models to generate pseudo-labels. The classification performance of untrained models is not very satisfactory. As a result, the quality of pseudo-labels  has no guarantee. Training models on poorly-labeled datasets will degrade the quality of these models. In contrast, in full SGL  which performs model pretraining in the first stage then using pretrained models to perform pseudo-labeling, the risk of generating poor-quality pseudo-labels is significantly reduced.

\begin{table}[t]
\caption{Results for ablation setting 1. ``With first stage" corresponds to the full SGL framework. ``Without first stage" corresponds to the formulation in Eq.(\ref{eq:ab-1}) where the first learning stage is removed. The results are classification errors on the test sets of CIFAR-10 and CIFAR-100.}
    \centering
    \begin{tabular}{l|c}
    \hline
    Method & Error (\%)\\
    \hline
    With first stage (CIFAR-100) &  
    \textbf{16.58}$\pm$0.18 \\
            Without first stage (CIFAR-100) & 17.33$\pm$0.20  \\
         \hline
              With first stage (CIFAR-10) & \textbf{2.41}$\pm$0.06  \\
         Without first stage (CIFAR-10) &2.60$\pm$0.05   \\
         \hline
    \end{tabular}
    \label{tab:ab1}
\end{table}

\begin{table}[t]
\caption{Results for ablation setting 2. ``Separate" means the first stage in SGL is conducted separately from the second stage. ``Joint" corresponds to the full SGL framework where the first and second stages are conducted jointly. 
    The results are classification errors on the test sets of CIFAR-10 and CIFAR-100.}
    \centering
    \begin{tabular}{l|c}
    \hline
    Method & Error (\%)\\
    \hline
    Joint (CIFAR-100) &  
    \textbf{16.58}$\pm$0.18 \\
           Separate (CIFAR-100) &  17.80$\pm$0.15 \\
         \hline
              Joint (CIFAR-10) & \textbf{2.41}$\pm$0.06  \\
          Separate (CIFAR-10) & 2.55$\pm$0.13\\
         \hline
    \end{tabular}
    \label{tab:ab2}
\end{table}

\begin{table}[t]
\caption{
    Results for ablation setting 3. ``With human labels" corresponds to the full SGL framework. ``Without human labels" corresponds to the formulation in Eq.(\ref{eq:ab-2}) where in the second stage each learner is trained only on pseudo-labeled datasets without using human-labeled datasets. The results are classification errors on the test sets of CIFAR-10 and CIFAR-100.
    }
    \centering
    \begin{tabular}{l|c}
    \hline
    Method & Error (\%)\\
    \hline
         With human labels (CIFAR-100) & 
          \textbf{16.58}$\pm$0.18 \\
          Without human labels (CIFAR-100) &16.96$\pm$0.17  \\
         \hline
          With human labels (CIFAR-10) & \textbf{2.41}$\pm$0.06  \\
         Without human labels (CIFAR-10) &2.74$\pm$0.15  \\
         \hline
    \end{tabular}
    \label{tab:ab4}
\end{table}

Table~\ref{tab:ab2} shows the classification errors on the test sets of CIFAR-10 and CIFAR-100, under the settings of ``Separate" and ``Joint" in the second ablation study. As can be seen, conducting the first and second stage separately leads to worse performance than performing them jointly, on both datasets. The reason is that: when performed separately, the second stage cannot influence the first stage; in the first stage, once the architectures are searched and network weights are trained, they will remain fixed. In contrast, in the full SGL framework where the first and second stage are performed jointly and share the same architecture, models trained in the second stage affect the validation loss, which subsequently affects the architecture; the changed architecture renders the models in the first stage to change as well. This can potentially bring in a synergistic effect: better models in the second stage results in better architectures in the third stage; better architectures improve the models in the first stage. Performing the first and second stage separately makes such a synergistic effect impossible to happen, which hence leads to inferior performance.

Table~\ref{tab:ab4} shows the classification errors on the test sets of CIFAR-10 and CIFAR-100 under the settings of ``with human labels" and ``without human labels" in the third ablation study. As can be seen, in the second stage of SGL, using only pseudo-labeled datasets for model training without leveraging human-labeled dataset yields worse performance. The reason is that  mutual training solely based on pseudo-labeled datasets  has a high risk of collective failure: if one model performs poorly, it will generate erroneous pseudo-labels, which renders other models trained using the pseudo-labels to fail as well. By leveraging an external human-labeled dataset where the labels are mostly correct, such a risk can be greatly reduced.

\begin{figure}[t]
    \centering
 \includegraphics[width=0.49\columnwidth]{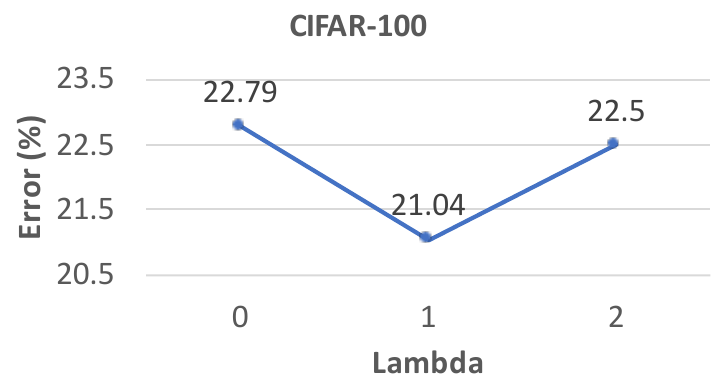}
  \includegraphics[width=0.49\columnwidth]{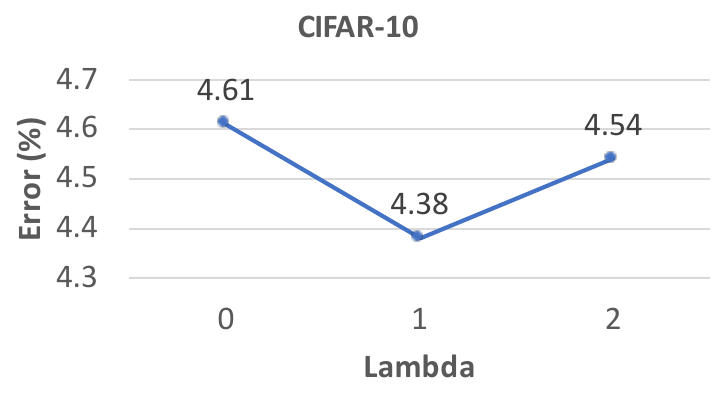}
       \caption{How classification errors change as the tradeoff parameter $\lambda$ increases. The errors are reported on the 5K randomly sampled data. 
       }
 \label{fig:lambda}
\end{figure}

\begin{figure}[t]
    \centering
 \includegraphics[width=0.49\columnwidth]{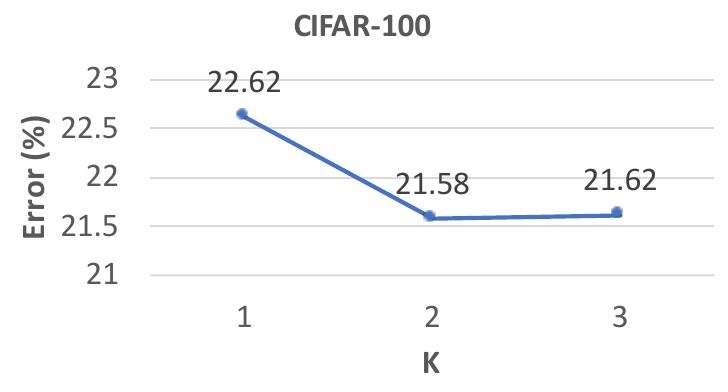}
  \includegraphics[width=0.49\columnwidth]{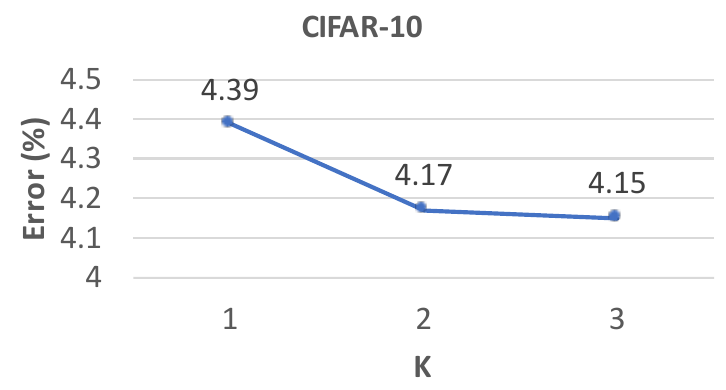}
       \caption{How classification errors change as the number of learners $K$ increases. The errors are reported on the 5K randomly sampled data. 
       }
 \label{fig:k}
\end{figure}

Figure~\ref{fig:lambda} shows the classification errors on CIFAR-100 and CIFAR-10 change as $\lambda$ increases. As can be seen, on CIFAR-100, when $\lambda$ increases from 0 to 1, the error decreases. This is because a larger $\lambda$ results in a stronger collaboration between learners where each learner actively relies on the pseudo-labeled datasets generated by other learners to train itself. A stronger collaboration enables different learners to help each other to improve. However, if $\lambda$ further increases to 2, the error increases. This is because too much emphasis is put on the pseudo-labeled datasets which are less reliable than the human-labeled training set. Ignoring the  human-labeled training set degrades the quality of trained models. Similar trend is observed in the CIFAR-10 results as well.

Figure~\ref{fig:k} shows how the classification errors on CIFAR-100 and CIFAR-10 vary as the number of learners $K$ increase from 1 to 3. As can be seen, when increasing $K$ from 1 to 2, the error decreases. Under $K=1$, there is no collaboration. When $K=2$, two learners collaboratively help each other to improve, hence achieving better performance. When $K$ increases from 2 to 3, the performance does not change significantly. This indicates that  two learners are sufficient for exploring the benefit of collaboration.

\section{Conclusions}
In this paper, drawing inspirations from the small-group learning (SGL) skill of humans, we propose a new ML framework to formalize SGL into a machine learning skill and leverage it to train better ML models. 
In SGL, a set of learners help each other to learn better: each learner uses its intermediately trained model to make predictions on unlabeled data examples and generates a pseudo-labeled dataset; meanwhile, each learner uses pseudo-labeled datasets generated by other learners to retrain and improve its model. To formalize SGL,  a multi-level optimization framework is proposed, which consists of three learning stages: each learner trains an initial model separately; all learners retrain their models collaboratively by mutually performing pseudo-labeling; all learners improve their neural architectures based on the feedback obtained during model  validation. We apply SGL to search neural architectures on CIFAR-100, CIFAR-10, and ImageNet. Experimental results demonstrate the effectiveness of our method.

\bibliography{release-2}

\end{document}